# Reconstruction of Compressed Spectral Imaging Based on Global Structure and Spectral Correlation

Pan Wang, Jie Li, Jieru Chen, Lin Wang, and Chun Qi, *Member, IEEE*

*Abstract*—In this paper, a convolutional sparse coding method based on global structure characteristics and spectral correlation is proposed for the reconstruction of compressive spectral images. The spectral data is regarded as the convolution sum of the convolution kernel and the corresponding coefficients, using the convolution kernel operates the global image information, preserving the structure information of the spectral image in the spatial dimension. To take full exploration of the constraints between spectra, the coefficients corresponding to the convolution kernel are constrained by the $L_{2,1}$ norm to improve spectral accuracy. And, to solve the problem that convolutional sparse coding is insensitive to low frequency, the global total-variation (TV) constraint is added to estimate the low-frequency components. It not only ensures the effective estimation of the low-frequency but also transforms the convolutional sparse coding into a de-noising process, which makes the reconstructing process simpler. Simulations show that compared with the current mainstream optimization methods, the proposed method can improve the reconstruction quality by up to 4 dB in PSNR and 10% in SSIM, and has a great improvement in the details of the reconstructed image.

## I. INTRODUCTION

Spectral image analysis plays an important role in environmental monitoring, geological exploration, art protection, life science research, and other fields [1]. Spectral image data is a 3-D data cube, including spatial and spectral information, which has a large volume. While an ordinary focal plane detector is usually a 2-D array, to obtain 3-D data cubes, spectral imaging instruments often need to scan the target in spatial or temporal dimensions, which almost loses the ability of temporal resolution. Compressed spectral imaging (CSI) [2-4] is based on compressed sensing technology [5,6], which greatly reduces the amount of sampling data of the detector, making it possible for the detector to obtain a 3-D data cube in a single integration time. The CSI essentially uses a 2-D detector to sample 3-D spectral data, so it needs to compress the third dimension, that is, the spectral dimension. This is achieved by multiplying images of each spectral segment by a 2-D coding mask with random patterns and finally, stacking along the spectral dimension to form a 2-D compressed coded observation image. This process is shown in Fig. 1.

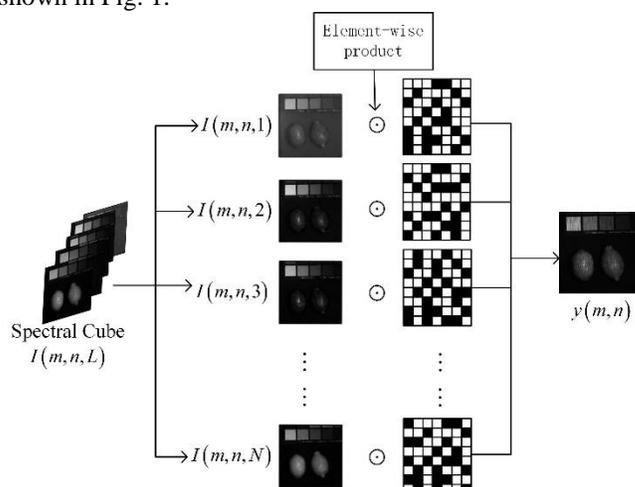

**Fig.1.** Schematic diagram of compressive spectral imaging

When the observed image is obtained, recovery is needed to get the original 3D spectral data, but the recovery of high-dimensional spectral data through a small amount of low-dimensional observed data is an underdetermined problem [7], so it is usually necessary to add additional priority constraints to solve it. For example, there is a lot of redundancy in the spectral image, which can be expressed sparsely in a specific transform domain, such as DCT or wavelet domain. Therefore, the transformed coefficients can be sparsely constrained by $L_1$ norm. Figueiredo et al. Proposed GPSR [8] method to solve $L_1$ norm, which has good generalization and usually takes less time to calculate. Yang et al. proposed to use the GMM [9,10] model to represent the spectral data by dividing the spectral image into multiple small patches. Then it is solved by Bayesian a posteriori probability estimation. However, when the image is divided into multiple small patches that will destroy the structure information. Similarly, when solving the $L_1$ constraint, the GPSR method usually converts the image into a 1-D vector for calculation and fails to make use of the structure information, resulting in limited improvement of the accuracy of this kind of method.

As mentioned above, the spectral image data is a 3-D data

This work was supported in part by the National Natural Science Foundation of China under Grant: 62275211, 61675161. *(Corresponding author: Jie Li).*

Pan Wang, Jie Li, Lin Wang, Jieru Chen, and Chun Qi are with the School of Information and Communications Engineering, Faculty of Electronic and Information Engineering, Xi'an Jiaotong University, Xi'an 710049, China. (e-mail: panwang@stu.xjtu.edu.cn; jielixjtu@xjtu.edu.cn; wangling92@stu.xjtu.edu.cn; jieruchen@stu.xjtu.edu.cn; qichun@mail.xjtu.edu.cn ).

Color versions of one or more of the figures in this article are available online at http://ieeexplore.ieee.org

cube, including 2-D spatial information and 1-D spectral information. Therefore, the spectral image is similar to the ordinary 2-D image in a specific spectral segment and has smoothness characteristics in the local region. Based on this assumption, Bioucas et al. proposed the TwIST algorithm [11] by introducing total variation [12] as a constraint. Wagadarikar et al. introduced the TwIST method into the coded aperture snapshot spectral imager (CASSI) [2-4] for spectral reconstruction. Yuan et al. proposed the GAP-TV [13] method. By transforming the general ADMM [14] framework into a solution that requires only a small number of iterative equations. However, the total variation constraint is the smoothing constraint on the image, which will lead to the loss of details in the reconstructed image. In recent years, some scholars have introduced the non-local self-similarity of images into image denoising[15] and have shown excellent results. Based on the non-local self-similarity, Liu et al. proposed the DeSCI [16] algorithm, which uses the low-rank characteristics of spatial and spectral dimensions for constrained reconstruction. The experimental results show an excellent effect [17]. However, the DeSCI method requires a large number of time-consuming 3D patches for block matching calculation, and block noise in reconstructed images affects the accuracy of image details.

With the wide application of deep learning, some scholars transform the compressed sensing reconstruction into a data-driven End-to-End mapping problem by training the deep CNN network[18-23], that is, inputting the 2D observation image and coding mask, outputting the 3D spectral data cube, which not only reduces the reconstruction time but also improves the reconstruction accuracy. While the End-to-End CNN network structure has a strong dependence on the structure of the observation system. When the system structure changes, such as changing the coding mask or sampling rate, the network needs to be retrained, resulting in poor generalization of this kind of method [17].

In conclusion, reconstruction methods for compressive spectral imaging systems not only need to ensure the accuracy of reconstruction but also need to be highly adaptable, which means the corresponding reconstruction accuracy should not change with the structure of the observation system. Therefore, a spectral reconstruction method based on convolutional sparse coding [24] is proposed in this paper. The spectral data is regarded as the convolution sum of multiple convolution kernels and corresponding sparse coefficients, which use the global convolution operations to ensure the image structure in the reconstructed results. Based on the properties of convolution of spectral data in [25], the response characteristics of spectral data on convolution kernel are analyzed, and the $L_{2,1}$ norm constraint is applied to the convolution coefficient, to make full use of the correlation between spectra to improve the reconstruction accuracy. Since the convolutional sparsity is usually insensitive to the low-frequency information in the image, this paper adds the global total-variation (TV) constraint to estimate the low-frequency part and also transforms the convolutional sparse coding into a denoising process to make the solution process simpler. To verify the effectiveness of the proposed algorithm in simulation experiments, the current mainstream optimization algorithms GAP-TV and DeSCI are compared. In terms of the reconstruction quality (PSNR, SSIM, and spectral accuracy), the proposed reconstruction method is superior to the current excellent DeSCI method and the GAP-TV algorithm. And for the reconstruction time, the proposed method is shorter than the DeSCI method. Compared with GAP-TV and DeSCI methods, the reconstructed images also have a great improvement in image detail. In addition, we compared various data-driven deep learning reconstruction algorithms, and the proposed algorithm still shows a good reconstruction effect.

This paper is organized as follows: In Section 2, the principle and structure of compressive spectral imaging are briefly introduced. In Subsection 3.1, the spectral reconstruction based on convolutional sparse coding is proposed and the correlation between spectral segments is analyzed. In Subsection 3.2, the influence of low-frequency components on convolutional sparse representation is analyzed, and a global TV constraint is added for iterative estimation of low-frequency components, which can improve the accurate estimation of low-frequency components and reduces the computational complexity. In Section 4, simulation experiments show that the proposed method can improve the reconstruction accuracy compared with the state-of-the-art methods.

## II. BRIEF DESCRIPTION OF THE CSI SYSTEM

Compressive spectral imaging (CSI) can sample high-dimensional spectral data at a lower Nyquist sampling rate and recover the original spectral data cube under constraint conditions, which is a computational spectral imaging technology. One of the more classic systems is coded aperture snapshot spectral imager (CASSI). The CASSI uses a 2-D mask in a special coding form to replace the slit in the traditional scanning spectral imager. The incident light first passes through the 2-D mask, then disperses through the dispersion element, and finally superimposed on the image sensor. This process is shown in Fig. 2.

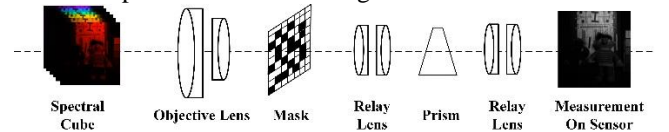

**Fig.2.** The optical structure of the classical CASSI system

Let $I \in R^{MNL}$ represent the spectral data cube, where $M, N$ represents the spatial dimension, $L$ is the spectral dimension, $H \in R^{MN \times MNL}$ is the observation matrix of the system, $\eta \in R^{MN}$ is the observation noise, and $y \in R^{MN}$ represents the vector form of 2-D observation data. Then, the imaging process of the CSI can be expressed as follows

$$y = HI + \eta \quad (1)$$

Since the dimension of the observation data $y$ is much smaller than the original spectral data $I$, solving (1) is an undetermined problem without a closed-form solution, which needs to be solved with the help of some prior knowledge. For example, spectral images are spatially and spectrally related which can be represented sparsely in a specific transform domain $\varphi$, such as DCT or Wavelet domain, thus the spectral data can be represented as $I = \varphi\theta$, and $\theta$ is the coefficient based on the transform domain $\varphi$. Therefore, the transformed sparse constraint can be used as a regular term for the coefficient $\theta$, as shown in (2).

$$\arg\min_{\theta} \frac{1}{2}\|y - H\varphi\theta\|_2^2 + \lambda\|\theta\|_1 \quad (2)$$

When solving (2), the alternating projection ADMM [14] method is usually used. However, to reduce the amount of data calculation, the image needs to be divided into blocks and converted into a one-dimensional vector for calculation, so the internal structure information of the image can not be fully utilized, which will affect the reconstruction quality.

### III. CONVOLUTIONAL SPARSE CODING FOR CSI

Convolutional sparsity [24] can be regarded as a special case of sparse representation [26]. It models the image as the convolution sum of a set of convolution kernels and corresponding coefficients. The convolution operation is usually done with the global image, which can keep the structural features of the image better. Its mathematical model is shown in (3), which $d_m$ represents the pre-learned convolution dictionary, and $x_m$ has the same dimension as the original signal $y \in R^{M \times N}$, which can be understood as the response coefficient of the original signal $y$ on the convolution kernel $d_m$. Typically, the distribution of $x_m$ is sparse that can be solved by $\|*\|_1$ constraints.

$$\arg\min_{\{x_m\}} \frac{1}{2}\left\|\sum_m d_m * x_m - y\right\|_2^2 + \lambda \sum_m \|x_m\|_1 \quad (3)$$

In the solving process, the ADMM method is usually used to introduce an auxiliary variable $\delta_m$ for solving, and the process is as follows: Firstly, (3) is transformed into:

$$\arg\min_{\{x_m\}} \frac{1}{2}\left\|\sum_m d_m * x_m - y\right\|_2^2 + \lambda \sum_m \|\delta_m\|_1 \quad (4)$$
$$s.t. \quad \delta_m - x_m = 0$$

Then, (4) is decomposed into the following parts:

$$x_m^{j+1} = \arg\min_{\{x_m\}} \frac{1}{2}\left\|\sum_m d_m * x_m - y\right\|_2^2 + \frac{\rho}{2}\sum_m \|x_m - \delta_m^j + u_m^j\|_2^2 \quad (5)$$

$$\delta_m^{j+1} = \arg\min_{\{\delta_m\}} \frac{1}{2}\|\delta_m\|_1 + \frac{\rho}{2}\sum_m \|x_m^{j+1} - \delta_m + u_m^j\|_2^2 \quad (6)$$

$$u_m^{j+1} = u_m^j + x_m^{j+1} - \delta_m^{j+1} \quad (7)$$

Equation (6) can be solved by the threshold shrinking method [24], while the convolution term in (5) cannot obtain a closed-form solution, and the convolution operation requires a large amount of computation. Therefore, (5) is usually transformed into the frequency domain and solved by the Sherman-Morrison equation [27]. The above content is the general process of convolutional sparse representation, which is usually used to process 2-D image signals. For 3-D spectral data, different representations are required. In the next section, convolutional sparse coding for 3-D spectral data will be introduced.

*A. Spectral reconstruction based on convolutional sparse coding*

Since the compressive spectral imaging system compresses the data of $L$ spectral segments into a single image $y \in R^{M \times N}$ through the observation matrix $H \in R^{MN \times MNL}$. It can be understood that each spectrum segment is computed element-wise with an encoding mask, as shown in Fig.1. To express this process clearly, the observed matrix $H$ is decomposed according to spectral segment with $H = \{h_1, h_2, \cdots, h_L\}$, where $h_j \in R^{MN \times MN}$ is the encoding mask corresponding to each spectrum segment. Finally adding the observation matrix $H = \{h_1, h_2, \cdots, h_L\}$ to (3), then it is transformed into

$$\arg\min_{\{x_{L,m}\}} \frac{1}{2}\left\|\sum_L h_L \sum_m d_m * x_{L,m} - y\right\|_2^2 + \lambda \sum_L \sum_m \|x_{L,m}\|_1 \quad (8)$$

$x_{L,m}$ can be regarded as the response coefficient of the convolution kernel $d_m$ on the 2-D image with the spectral segment $L$, $\lambda$ is the parameter that controls the importance of the regularization term. In (8), the spatial dimension of the spectral image is constrained globally by the $\|*\|_1$ constraint on the convolution coefficient $x_{L,m}$. However, constraints on spatial dimensions alone is not enough to improve the reconstruction accuracy of the spectral dimension, so it is necessary to consider adding prior knowledge of spectral dimensions to improve the accuracy of reconstructed data. From the spectral dimension, there is little change in the image structure between different spectral segments, only the light and dark changes at the same position. Therefore, when using the same convolution kernel $d_m$ to convolve the different spectral images, there is only a small difference in the coefficient $x_{L,m}$ theoretically. This can be used as a constraint to improve spectral accuracy. To verify the above conclusions, this paper analyzed the coefficient graphs corresponding to different convolution kernels $d_m$, and the results are shown in Fig. 3.

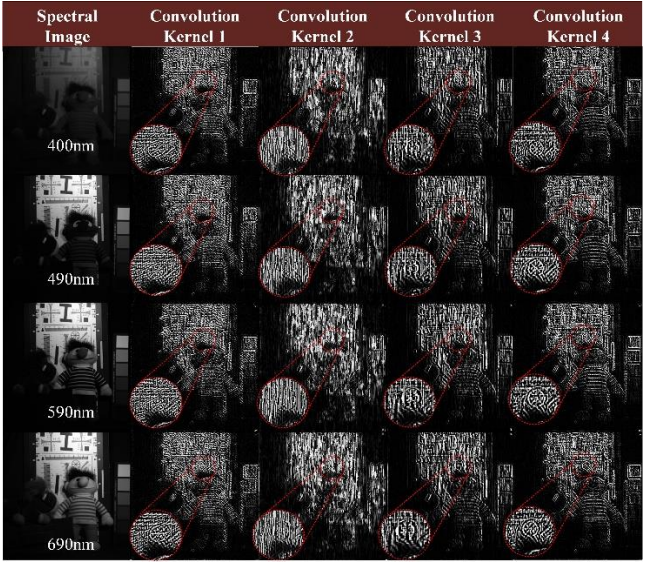

**Fig.3.** Response coefficient corresponding to convolution kernel(The first column contains raw spectral data for different spectrum segments, and The second to fifth columns are the images composed of the response coefficients of the original spectral image on four different convolution kernels)

### B. Explore the constraints between spectra

The spectral data in Fig. 3 is from the CAVE dataset [28], with a size of $512 \times 512$ in spatial dimensions, and a size of 31 bands in spectral dimensions with a spectral range of 400nm-700nm, and each convolution core size is $12 \times 12$. It can be intuitively found in Fig.3, that for the same spectral segment, the results of different convolution kernels are quite different. And for the different spectral segments, although the image differences are large, the response differences corresponding to the same convolution kernel remain very small. This can be understood as the response of 2-D spectral images of different spectral segments on the same convolution kernel is stable. Therefore, this characteristic is introduced into the spectral reconstruction constraint term as $\|*\|_{2,1}$ to constrain the spectral data, that is, $\|*\|_2$ constraint is used across spectral segments, while $\|*\|_1$ is still used for sparse constraint in each spectral segment. Therefore, the reconstruction constraint equation is modified as

$$\arg\min_{\{x_{L,m}\}} \frac{1}{2}\left\|\sum_L h_L \sum_m d_m * x_{L,m} - y\right\|_2^2 + \rho \left\|\{x_{L,m}\}\right\|_{2,1} \quad (9)$$

Where $\rho$ is the regularization parameter that controls the importance of the regularization term. In the solution process, since convolution and observation matrix $H$ exist simultaneously in (9), it cannot be transformed to the frequency domain for a simplified solution. Therefore, the solution method proposed by Heide et al. [29] is adopted to solve this problem containing by summing up multiple subproblems. To verify the reconstruction effect, simulation experiments were carried out here. The SSCSI system proposed in [30] was adopted in the simulated spectral imaging device, which realized simultaneous coding of spatial and spectral dimensions by using only a group of dispersion elements, and it is an efficient compressed sensing spectral acquisition system [31]. The experimental data still use the CAVE dataset in [28]. For the selection of the convolution dictionary, referred to the conclusion in [32], that is, the convolution dictionary trained with data similar to the content to be reconstructed has little difference from the general dictionary in the accuracy of reconstruction results. Therefore, without loss of generality, this paper also uses the general dictionary in [29] (the corresponding size $12 \times 12 \times 144$) for reconstruction, and the results are shown in Fig.4.

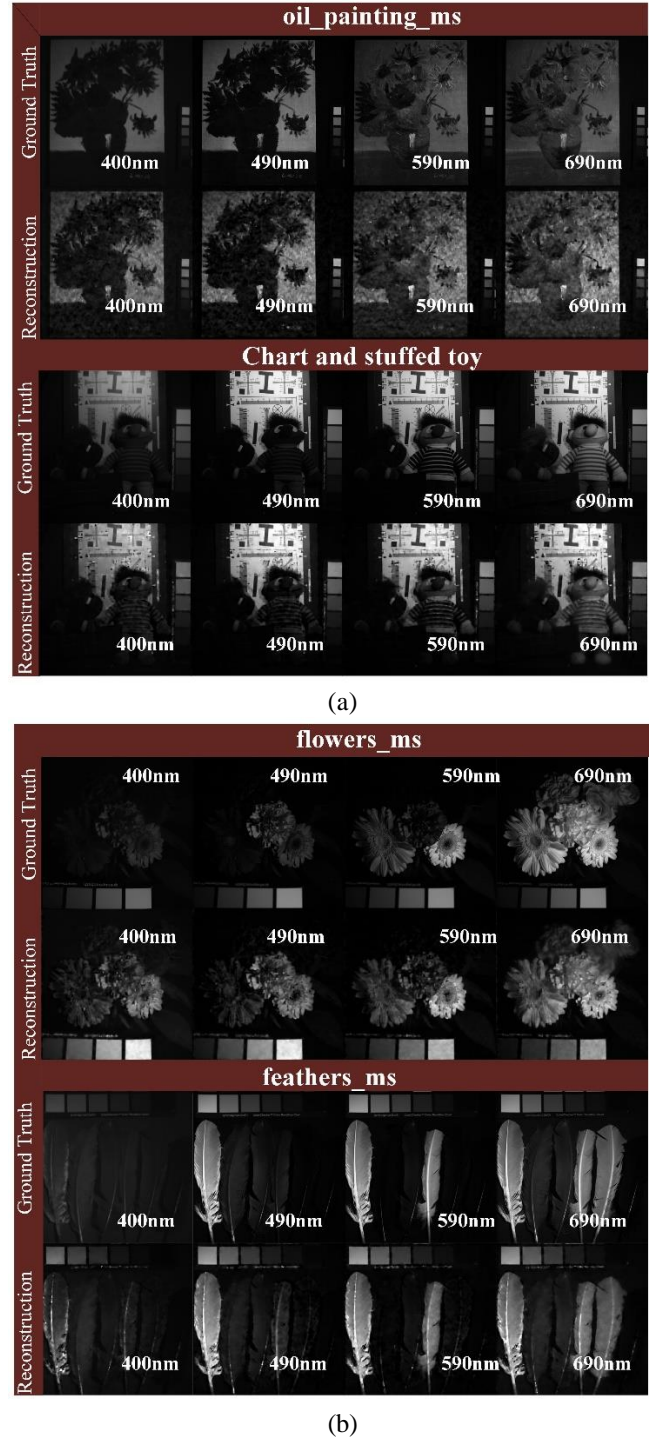

**Fig.4.** Convolutional sparse representation of spectral reconstruction results

It can be seen from the reconstruction results in Fig. 4 that the reconstruction results in the spectral dimension can better reflect the changing trend of the ground truth data through the constraint of the $\|*\|_{2,1}$ norm, but for the spatial dimension, there is a large block noise. This is due to the constraint of convolution coefficients by $\|*\|_1$ norm in spatial dimension is not enough to suppress the noise in reconstructed images. Therefore, the next sub-section will consider improving the reconstruction quality of the spatial dimension.

*C. Estimation of low-frequency components in convolutional sparse*

According to the conclusion in [25], the convolutional sparse coding is usually sensitive to the high-frequency part of the image, while cannot effectively represent the low-frequency component. The results in Fig.3 also reflect this characteristic, therefore it is necessary to add additional estimation for the low-frequency components. For example, in [25] the low-pass filter is used to estimate the low-frequency component from the observation directly. But in the compressive spectral imaging system, such as CASSI or SSCSI, affected by encoding mask, the obtained observation image will have a large number of coding superpositions, which have a great impact on low-frequency component estimation, as shown in Fig. 5.

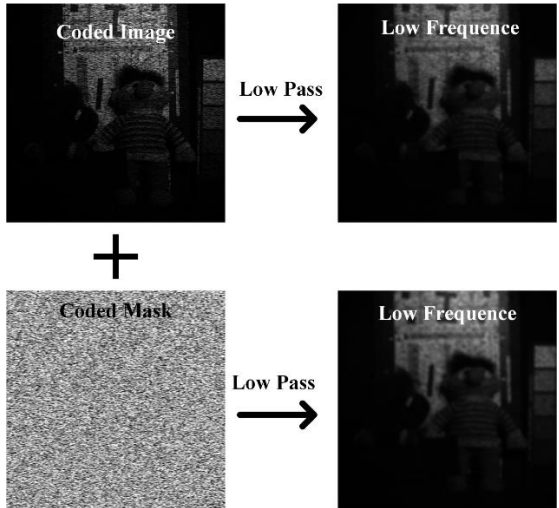

**Fig.5.** Estimate the low-frequency component from the observed image(The first line uses low-pass filtering to obtain the low frequency directly from the encoded observation image, and The second line is combined with the low-pass filtering of the coding mask to estimate the low-frequency part)

In Fig. 5, after coding by the mask, it can be seen that there are a large number of coding patterns of different sizes in the coded image, which are difficult to remove. In [32] Serrano et al. proposed combining the coding mask prediction and interpolation filtering to solve this problem. Despite a certain level of improvement over the direct use of low-pass filtering, it still can not eliminate the effect of the coded mask. Therefore, how to estimate the low-frequency part of the reconstructed spectral image is a key point to improve the reconstruction quality.

It can be seen from the above content, that due to the insensitivity of convolutional sparse representation to low frequency, a large number of noises in the reconstruction results cannot be eliminated, so additional denoising processing is needed. As mentioned above, the total variation (TV) constraint shows a good effect when smoothing the spatial dimension, so this paper proposes to use the TV constraint to update the estimation of low-frequency components in the iterative process, and modify the reconstruction constraint equation as follows:

$$\arg\min_{\{x_{L,m}\}} \frac{1}{2}\left\|\sum_L h_L \sum_m d_m * x_{L,m} - y\right\|_2^2 + \beta \sum_L TV(\sum_m d_m * x_{L,m}) + \rho\left\|\{x_{L,m}\}\right\|_{2,1} \quad (10)$$

In (10), TV constraints are imposed on the spatial dimension of each spectral segment, and the low-frequency estimation problem is transformed into an optimization problem, which can be solved together in the reconstruction process. Similar to (9) in the solution process, convolution $*$ and observation matrix $H$ also exist in (10), which cannot be transformed to the frequency domain for a simplified solution. In this paper, (10) is divided into two parts for the solution, as follows:

$$\begin{cases} \arg\min_{I} \frac{1}{2}\|HI - y\|_2^2 + \beta \sum_L TV(I_L) \\ \arg\min_{\{x_{L,m}\}} \frac{1}{2}\sum_L\left\|\sum_m d_m * x_{L,m} - I_L\right\|_2^2 + \rho\left\|\{x_{L,m}\}\right\|_{2,1} \\ s.t. \quad I_L = \sum_m d_m * x_{L,m} \end{cases} \quad (11)$$

In (11), the observation process is calculated separately from the convolution process. Firstly, the optimization based on TV constraint is used to make a preliminary estimate of the spectral data $I$. Then the convolutional sparse representation of each band of spectral data $I_L$ is performed with the $\|*\|_{2,1}$ as the constraint to improve reconstruction accuracy. The convolutional sparse deconvolution problem with the observation matrix is simplified to a convolutional sparse denoising problem to reduce the complexity. Therefore, the solution steps corresponding to (11) are as follows:

Firstly, the optimization equation of TV constraint is solved by the GAP method [13], and the auxiliary variable $v$ is introduced:

$$\begin{cases} \arg\min_{I} \|y - HI\|_2^2 + \|v - I\|_2^2 \\ \arg\min_{v} \|v - I\|_2^2 + \beta TV(v) \\ s.t. \quad I = v \end{cases} \quad (12)$$

And the solution corresponding to (12) is:

$$\begin{cases} v^{(t)} = v^{(t-1)} + H^T(HH^T)^{-1}(y - HI^{(t-1)}) \\ I^{(t)} = \text{TV}(v^{(t)}) \end{cases} \quad (13)$$

After obtaining the spectral data cube $I$ in the (13), the convolutional sparse representation is performed to further improve reconstruction accuracy. Since the observation matrix $H$ is not included, it can be solved by using the general Sherman-Morrison equation combined with the threshold shrinkage method. The alternative solution process can refer to (5) to (7). Therefore, the solution process of the reconstruction model proposed in this paper can be summarized as follows:

---

Algorithm 1: The main process of the algorithm in this paper

1: **for** t = 0 to Max-iter, **do**
2:   $I^{t-1}$: Use Eq. (13) to get $I^{t-1}$
3:   **for** j = 0 to Max-iter, **do**
4:     update $\{x_{L,m}\}$:

$$\{x_{L,m}^{j+1}\} = \arg\min_{\{x_{L,m}\}} \frac{1}{2} \sum_L \left\| \sum_m d_m * x_{L,m} - I_L^{t-1} \right\|_2^2 + \frac{\rho}{2} \sum_L \sum_m \left\| x_{L,m} - \delta_{L,m}^j + u_{L,m}^j \right\|_2^2$$

5:     update $\{\delta_{L,m}\}$:

$$\delta_{L,m}^{j+1} = \arg\min_{\{\delta_{L,m}\}} \frac{1}{2} \left\| \delta_{L,m} \right\|_{2,1} + \frac{\rho}{2} \sum_m \left\| x_{L,m}^{j+1} - \delta_{L,m} + u_{L,m}^j \right\|_2$$

6:     update $\{u_{L,m}\}$: $u_{L,m}^{j+1} = u_{L,m}^j + x_{L,m}^{j+1} - \delta_{L,m}^{j+1}$
7:   **end for**

$$I^t = \sum_L \sum_m d_m * x_{L,m}$$

8: **end for**

**Output: hyperspectral images $I$.**

---

## IV. EXPERIMENTS AND ANALYSIS

In this section, the reconstruction method proposed in this paper is verified through simulation experiments. The simulated compressive spectral imaging system is SSCSI. The experimental data adopts the CAVE spectral data in [28] with a size of $512 \times 512 \times 31$ and a spectral range of 400nm - 700nm. The convolution kernels used in the experiment adopt the general convolution kernels in [29], with a size of $12 \times 12 \times 144$.

### A. Ablation experiments for Low-frequency estimation

Firstly, to verify the effectiveness of convolutional sparse reconstruction combined with TV constraint for low-frequency estimation, we performed an ablation experiment. The reconstruction method using only convolutional sparse representation is compared. The experimental results are shown in Fig.7

As can be seen from Fig. 7, when only the convolutional sparse coding(CSC without TV) is used for reconstruction, the "coding noise" cannot be eliminated because the low-frequency part is estimated from the coded image directly, so there is a large error in some areas of the results. As a comparison, the method proposed in this paper can accurately estimate the low-frequency part of the spectral image due to the update of the estimation based on TV constraints.

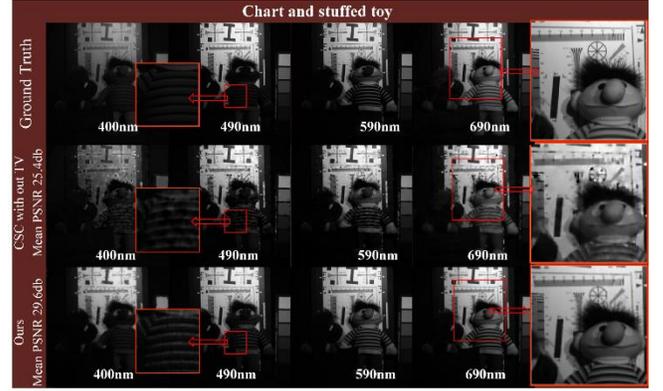

**Fig.7.** Influence of low-frequency estimation on reconstruction results

### B. Comparisons With State-of-the-Arts

Advanced compressed sensing reconstruction methods such as GAP-TV [13] and DeSCI[16] are compared to illustrate the reconstruction accuracy of the proposed algorithm. The peak signal-to-noise ratio (PSNR) and structural similarity (SSIM) are compared in spatial dimensions. In terms of spectral accuracy, specific image blocks are selected to compare the consistency between the reconstructed results and the ground truth distribution curve of each band. The results are shown in Fig.8.

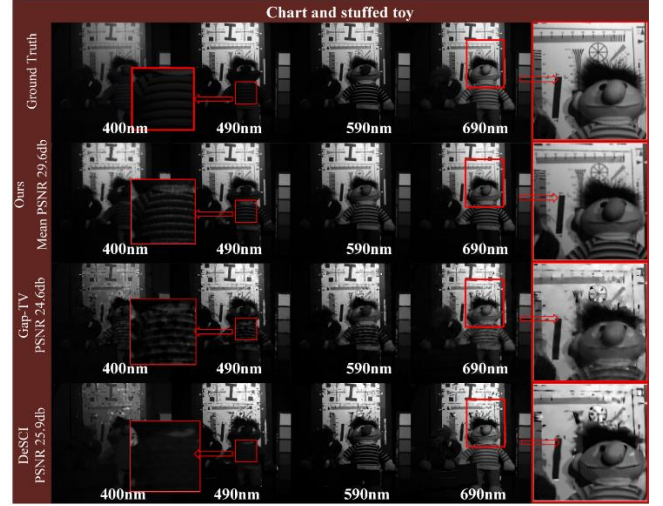

**Fig.8.** Quality comparison of spatial dimension(The first line is the ground truth spectral data. The second line is the recovery result of the algorithm in this paper, and the average PSNR is 29.6 dB. The third row is the reconstruction result of the GAP-TV algorithm, and the average PSNR is 24.6 dB. The fourth row is the reconstruction result of the DeSCI algorithm, and the average PSNR is 25.9 dB. The red zoom box shows the quality of the details of the recovery result.)

As can be seen in Fig.8, compared with GAP-TV and DeSCI methods, the proposed method achieves higher reconstruction quality in the spatial dimension, and the original data can be reconstructed more accurately in local details. To compare the spectral accuracy, four image blocks are selected to draw the normalized intensity corresponding to each reconstruction algorithm from 400 nm to 700 nm. The results are shown in Fig.9.

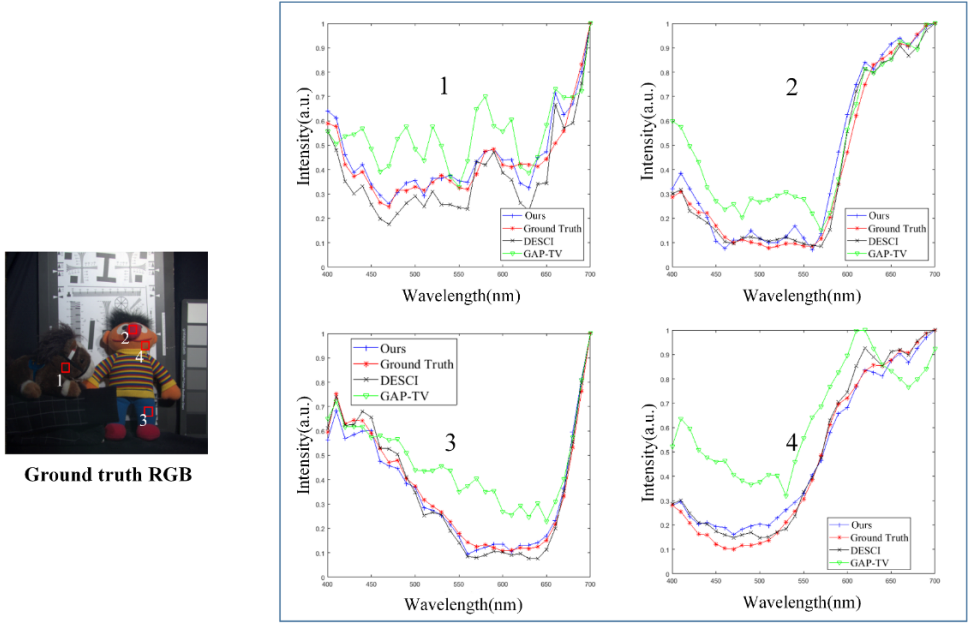

Fig.9. Comparison of spectral consistency of reconstructed data from different algorithms(The image on the left is the RGB image corresponding to the spectral data, in which 4 small blocks are selected as areas for comparison of spectral accuracy. The four sub-images on the right correspond to the spectral distribution curves in the four small blocks on the left)

In Fig. 9, the spectral distribution curve of the proposed algorithm is similar to the currently recognized cutting-edge algorithm DeSCI which is close to the ground truth and has obvious advantages over the GAP-TV method. To quantify the accuracy, spectral angle mapping (SAM) is used as a comparison. In Fig.8, the four regions on the left are selected for average SAM statistics, and the corresponding results are shown in Table 1.

TABLE I
SPECTRAL ACCURACY COMPARISON SAM （RAD）

| Method | Region 1 | Region 2 | Region 3 | Region 4 |
|---|---|---|---|---|
| Gap-TV | 0.7316 | 0.6041 | 0.6857 | 0.9434 |
| DeSCI | 0.5449 | 0.2643 | 0.0800 | **0.1305** |
| Ours | **0.4628** | **0.2615** | **0.0749** | 0.1973 |

It can be seen from Table 1, each algorithm calculates the SAM with the ground truth. The SAM angle of our algorithm and DeSCI algorithm is small (The smaller the SAM value, the higher the accuracy), which has a great improvement in spectral accuracy than GAP-TV algorithm.

### C. Multi-scene data experiment

To verify the adaptability of the method in this paper on different data sets, a total of 10 scenarios[28] are selected for testing, as shown in Fig 10.

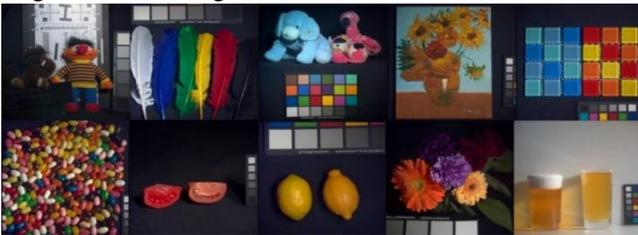

**Fig.10.** Dataset images of 10 scenes (The first row from left to right are toystuff, weather, toys, oil, and glass. The second row from left to right are beans, tomato, lemon, flower, and beer) The data size is $512 \times 512 \times 31$.

The reconstruction quality of the proposed algorithm is compared with that of GAP-TV and DeSCI in 10 scenarios. To further compare the reconstruction methods based on deep learning, we selected two mainstream deep learning networks (Unet[33] and ADMM-net[34]) to compare the reconstruction results. The measurement indexes are mean peak signal-to-noise ratio (PSNR) and mean structural similarity (SSIM), and the corresponding results are shown in Table 2 and Table 3.

TABLE 2
RECONSTRUCTION QUALITY COMPARISON 1

| Method | Metric | Toystuff | Oil | Flower | Feather | Toys |
|---|---|---|---|---|---|---|
| Gap-TV | PSNR(dB) | 24.6 | 22.46 | 19.37 | 19.49 | 24.98 |
|  | SSIM | 0.8606 | 0.7425 | 0.8218 | 0.7906 | 0.8447 |
| DeSCI | PSNR(dB) | 25.91 | 24.94 | 22.27 | 24.74 | 26.13 |
|  | SSIM | 0.9094 | 0.7535 | 0.8554 | 0.8759 | 0.9138 |
| Unet | PSNR(dB) | 27.18 | 26.13 | 28.56 | 24.86 | 22.32 |
|  | SSIM | 0.9155 | 0.8662 | 0.8566 | 0.7924 | 0.7276 |
| ADMM net | PSNR(dB) | **31.03** | **28.66** | **30.55** | 28.18 | 24.11 |
|  | SSIM | **0.9368** | **0.9458** | **0.9488** | 0.8688 | 0.8956 |
| Ours | PSNR(dB) | 29.6 | 27.33 | 26.38 | **28.24** | **28.00** |
|  | SSIM | 0.9222 | 0.7914 | 0.8957 | **0.8826** | **0.9195** |

TABLE 3
RECONSTRUCTION QUALITY COMPARISON 2

| Method | Metric | Glass | Bean | Tomato | Lemon | Beer |
|---|---|---|---|---|---|---|
| Gap-TV | PSNR(dB) | 21.78 | 23.37 | 26.52 | 23.64 | 27.17 |
|  | SSIM | 0.7276 | 0.7825 | 0.9309 | 0.8925 | 0.8623 |
| DeSCI | PSNR(dB) | 23.98 | 23.45 | 28.19 | 30.63 | 34.06 |
|  | SSIM | 0.6021 | 0.7307 | 0.9410 | 0.9362 | **0.9622** |
| Unet | PSNR(dB) | 22.50 | 20.84 | 33.86 | 30.42 | 25.64 |
|  | SSIM | 0.7568 | 0.7805 | 0.9260 | 0.8986 | 0.8279 |
| ADMM net | PSNR(dB) | **26.59** | 21.89 | **36.67** | **35.09** | 29.58 |
|  | SSIM | **0.8862** | 0.7982 | 0.9268 | 0.9488 | 0.8650 |
| Ours | PSNR(dB) | 25.23 | **26.75** | 30.06 | 32.27 | 34.46 |
|  | SSIM | 0.8070 | **0.8118** | 0.9486 | 0.9411 | 0.9559 |

It can be seen from the results in Table 2 and Table 3 that the reconstruction results of the proposed method in this paper show good accuracy in multiple scenes. The ADMMnet network based on the deep learning method indeed shows a high reconstruction accuracy, which is competitive with the proposed algorithm in our article. To compare the timeliness of reconstruction, a statistical comparison is made on the reconstruction time. The computer configuration used in all experiments is i7-7700K CPU 4.20 GHz, 16 GB RAM, NVIDIA GTX1080Ti GPU. Among them, the method based on deep learning shows excellent real-time performance, and the reconstruction time is within a few seconds to tens of seconds. Inside the optimization method, GAP-TV takes 204 seconds, showing excellent timeliness, while the DeSCI method takes more time, up to 26 hours, due to the need for a large number of block matching algorithms, the reconstruction time of the proposed method is about 50 minutes, which greatly reduces the reconstruction time while ensuring high-quality reconstruction results.

## V. Conclusion

In compressive spectral imaging, the spectral data cube corresponding to the scene can usually be recovered from the 2D observation image with a low sampling rate. In this paper, sparse convolutional coding technology is used to reconstruct the 3D spectral data. By treating the spectral data as the convolution-weighted sum of several convolution kernels and corresponding sparse coefficients. Thus, the image structure information can be better preserved in the reconstruction results. After analysis, when the same convolution kernel is applied to different spectral segments of spectral data, the corresponding response coefficient is only a small difference in magnitude. To maximize the use of this feature, the coefficient corresponding to the convolution kernel is constrained by $\|*\|_{2,1}$ norm to improve the reconstruction accuracy. Given the insensitivity of convolutional sparse coding to low-frequency information, this paper adds the global total variation (TV) constraint. It not only ensures the effective estimation of the low-frequency part of the spectral image but also converts the sparse convolutional coding into a denoising process, which reduces the computational complexity and simplifies the solution process. The simulation results show that the method proposed in this paper has a great improvement in the details of the reconstructed image compared with the current mainstream methods. For further research work, we plan to introduce a 3-D convolution kernel to replace the 2-D convolution kernel and consider introducing a multi-scale dictionary to improve the precision of reconstruction and reduce the calculation amount using deep unfolding technology.

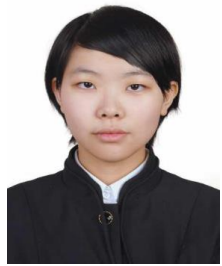

**JieRu Chen** received the B.S degrees from the school of Electronic Engineering, Xidian University, Xi'an, Shaanxi, China, in 2018. Received the M.S degrees from the school of Communication Engineering, Xi'an Jiaotong University, Xi'an, Shaanxi, China, 2021. Her main research interests include image denoising, compressive sensing.

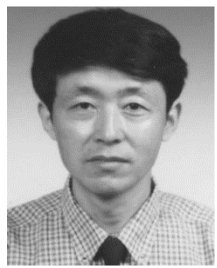

**Chun Qi** received the Ph.D. degree from Xi'an Jiaotong University, Xi'an, China, in 2000. He is currently a Professor and a Ph.D. Supervisor with the School of Information and Communications Engineering, Xi'an Jiaotong University. His current research interests mainly include image processing, pattern recognition, and signal processing.

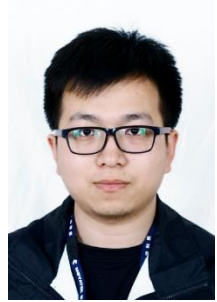

**Pan Wang** received the B.S. and M.S. degrees from the School of Automation and Information Engineering, Xi'an University of Technology, Xi'an, Shaanxi, China, in 2012 and 2015, respectively. He is currently working toward the Ph.D. degree at Xi'an Jiaotong University, Xi'an, Shaanxi, China. His current research interests mainly compressed spectral imaging and Computational imaging

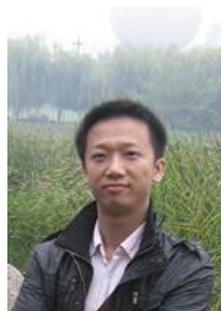

**Jie Li** completed his BS degree in Physics at Xi'an Jiaotong University in Xi'an, China in 2006. Upon graduation, his personal interest in Spectral Imaging and Polarization Imaging lead him to obtain his MS and PhD in Optical Sciences and Electronic Science and Technology at Xi'an Jiaotong University, in 2009 and 2012, respectively. Now, he is a Professor at the School of Information and Comunications Engineering, Xi'an Jiaotong University. Research performed at the school included visible and infrared imaging spectrometry, polarimetry, and interferometry.

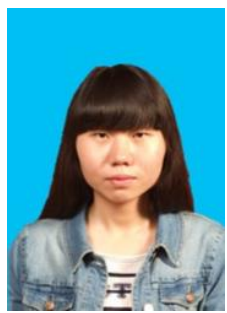

**Lin Wang** received the Ph.D. degree from Xi'an Jiaotong University, Xi'an, China, in 2000. He is currently a Professor and a Ph.D. Supervisor with the School of Information and Communications Engineering, Xi'an Jiaotong University. His current research interests mainly include image processing, tion, and signal processing.